\documentclass[final,5p]{elsarticle}

\usepackage{amsmath}
\usepackage{amssymb}
\usepackage{times}
\usepackage{multirow}
\usepackage{color}
\usepackage{graphicx}
\usepackage{subfigure}
\usepackage{url}
\usepackage{algorithm}
\usepackage{algorithmic}

\usepackage{ifpdf}
\ifpdf   % We're running pdfTeX in PDF mode
  \graphicspath{{./pdf/}{./figures/}}
   \DeclareGraphicsExtensions{.pdf,.jpeg,.png,.jpg}
\else    % We're not running pdfTeX, or running pdfTeX in DVI mode
  \graphicspath{{./eps/}}
  \DeclareGraphicsExtensions{.eps,.ps}
\fi

\usepackage{hyperref}
\hypersetup{bookmarks=true,
    bookmarksdepth=3,
    bookmarksopen,
    bookmarksnumbered,
    pdfstartview=FitH,
    colorlinks=true,
    breaklinks=true,
}

\usepackage[pagewise]{lineno}
\modulolinenumbers[1]

\def\mm#1{#1}

\journal{Image and Vision Computing}

\begin{document}

\begin{frontmatter}

\title{Learning Deep Similarity Models with Focus Ranking for Fabric Image Retrieval}

%\author{Daiguo Deng  \and Ruomei Wang \and Hefeng Wu \and Huayong He \and Xiaonan Luo}
%\address{XXX}

\author[label1]{Daiguo Deng}
\author[label1]{Ruomei Wang}
\author[label2]{Hefeng~Wu\corref{mycorrespondingauthor}}
\cortext[mycorrespondingauthor]{Corresponding author.}
\ead{wuhefeng@gmail.com}
\author[label1]{Huayong He}
\author[label4]{Qi Li}
\author[label3]{Xiaonan Luo}

\address[label1]{Sun Yat-sen University, Guangzhou 510006, China}
\address[label2]{Guangdong University of Foreign Studies, Guangzhou 510006, China}
\address[label4]{Western Kentucky University, Bowling Green, KY, 42101, USA}
\address[label3]{Guilin University of Electronic Technology, Guilin 541004, China}

\begin{abstract}
Fabric image retrieval is beneficial to many applications including clothing searching, online shopping and cloth modeling. Learning pairwise image similarity is of great importance to an image retrieval task. With the resurgence of Convolutional Neural Networks (CNNs), recent works have achieved significant progresses via deep representation learning with metric embedding, which drives similar examples close to each other in a feature space, and dissimilar ones apart from each other. In this paper, we propose a novel embedding method termed {\it focus ranking} that can be easily unified into a CNN for jointly learning image representations and metrics in the context of fine-grained fabric image retrieval. Focus ranking aims to rank similar examples higher than all dissimilar ones by penalizing ranking disorders via the minimization of the overall cost attributed to similar samples being ranked below dissimilar ones. At the training stage, training samples are organized into focus ranking units for efficient optimization. We build a large-scale fabric image retrieval dataset (FIRD) with \mm{about 25,000 images of 4,300 fabrics}, and test the proposed model on the FIRD dataset. Experimental results show the superiority of the proposed model over existing metric embedding models.

\end{abstract}


\end{frontmatter}

%\linenumbers

\section{Introduction}

Fabric image retrieval, as a special case of generic image retrieval, benefits a wide range of applications, e.g., large-scale clothing searching, online shopping and cloth modeling. Given a query fabric image, a retrieval algorithm is expected to output fabric images that are identical or similar to the query image. It remains a challenging task
due to severe variations in illuminations, orientations, scales and wrinkles among fabric images.

Traditional methods for image retrieval mainly involve two critical components: i) design a robust and discriminative
image representation, and ii) determine an effective distance or similarity metric for a given image representation.
Image representations used in a traditional method are usually hand-crafted, e.g., SIFT \cite{lowe2004distinctive},
GIST \cite{oliva2001modeling,douze2009evaluation}, Bag of Words (BoW) \cite{sivic2003video}, Fisher
Vector (FV) \cite{perronnin2007fisher,perronnin2010improving}, and VLAD \cite{jegou2010aggregating}.
Although achieving reasonable successes in image retrieval, these methods depend heavily on feature engineering.
More seriously, the two components are designed or learned separately, leading to a sub-optimal solution.

Recently, methods were proposed to learn an image representation and a distance or similarity metric jointly
\cite{wang2014learning,hoffer2015deep} based on Convolutional Neural Networks (CNN)
\cite{krizhevsky2012imagenet,zeiler2014visualizing}, which can be used seamlessly in image retrieval.
Specifically, these methods trained a CNN with metric learning embedding. Two simple yet effective metric learning embedding methods are pair \cite{hadsell2006dimensionality,chopra2005learning} and triplet embedding \cite{schroff2015facenet,weinberger2009distance}.  These two embedding methods are optimized to pull samples of different labels apart from each other and push samples with the same labels close to each other. The most important advantage of these discriminative models is that they can jointly learn an image representation and semantically meaningful metric, which is more robust against intra-class variations and inter-class confusions.

An image retrieval system aims to find out samples with the same labels against a great many negative ones. But pair and triplet embedding methods model a metric with no more than one negative image as reference, which are extremely rough approximation to a real setting. In this paper, we propose a novel embedding method named {\it focus ranking}, which can easily be unified into a CNN for a joint optimization. In particular, the proposed model aims to rank a sample with the same label (i.e., the matched sample) top over all negative ones. Hence, we penalize any ranking disorder by minimizing the overall cost of the matched samples ranked below any negative ones. At the training stage, we organize training samples into focus ranking units, each of which consists of a probe sample, a matched sample, and a reference set, for an efficient optimization. It learns to rank the matched sample top over all the negative ones in the reference set.

To the best of our knowledge, there are not fabric image retrieval datasets publicly available. We build a large-scale fabric image retrieval datasets (FIRD). \mm{It contains 4,300 fabrics, each of which has five to ten instances}. We divide the FIRD dataset into two partitions, via randomly selecting half of the fabrics as the training set and the rest as the test set. We randomly select 2/5 images as queries for each fabric in the test set and the rest form the retrieval set. Given a query image, we aim to find out the ones of the same fabric at the retrieval set. Extensive experiments on FIRD demonstrate that the proposed model outperforms existing ones by a large margin.

In summary, this paper makes four main contributions to the community.

\begin{itemize}
\item  We propose a novel focus ranking embedding method, which aims to rank the matched sample top over any negative ones. This method can be easily unified into a CNN to learn a deep similarity model by optimizing the image representation and metric jointly. Compared to existing embedding methods, \mm{it can learn more robust and discriminative representation to reduce intra-class variations and enhance inter-class discrimination}.

\item We apply focus ranking unit generation to model training, which highlights the focus by the negative-positive ratio in the process of training. It can also help increase the diversity of training samples, which is greatly important to learning.

\item We build a large-scale fabric image retrieval dataset (FIRD), which contains \mm{4,300 fabrics and about 25,000 images} for training and test. To the best of our knowledge, it is the first large-scale dataset for this task.% (4000$\times$5$\times$2)

\item Extensive experiments demonstrate the superiority of the proposed method over existing metric embedding methods. We also carefully evaluate and discuss key components of our model that improve the performance.
\end{itemize}

The rest of this paper is organized as follows. Related works are reviewed in Section \ref{sec:related_work}. The proposed model is presented in Section \ref{sec:deeprankingmodel}, and the FIRD dataset is described in Section \ref{sec:dataset}. Section \ref{sec:experiment} presents an extensive evaluation and comparison between the proposed method and several state-of-the-art methods, in addition to the component analysis of the proposed method. Section \ref{sec:conclusion} concludes the paper.

\section{Related Works}
\label{sec:related_work}
In this section, we review three topics most related to this work according to its objective and the technical components.
These three topics are traditional descriptors, deep representation learning and distance learning.

\subsection{Traditional descriptors}
Traditional image retrieval systems mainly focus on designing discriminative hand-crafted image representations for matching two images. Scale-Invariant Feature Transform (SIFT) \cite{lowe2004distinctive} was used widely in image retrieval, due to its distinct invariance to image translation, scaling, and rotation. Many works applied Bag of Features (BOF) using SIFT features for large scale image retrieval, such as \cite{sivic2003video}. They first quantized local feature vectors into visual words, and represented the image by the frequency histogram of visual words. Nister and Stewenius \cite{Nister2006SRV}
hierarchically quantized local descriptors to a hierarchical vocabulary, which improves not only the retrieval efficiency
but also its quality. Jegou et al. \cite{jegou2007contextual} presented a contextual
dissimilarity measure for accurate and efficient image search, while Fraundorfer et al. \cite{fraundorfer2007binning} proposed a binning scheme for fast hard drive based image search. Jegou et al. \cite{jegou2008hamming, jegou2010improving} developed a more precise representation by integrating hamming embedding and weak geometric consistency within the inverted file. Some works also aggregated a set of local descriptors into global ones for large scale image retrieval, e.g., Fisher Vector (FV) \cite{perronnin2007fisher, perronnin2010improving}, and Vector of Locally Aggregated Descriptors (VLAD) \cite{jegou2010aggregating}. Perronnin et al. \cite{perronnin2010large} represented an image as a Fisher Vector, and compressed Fisher Vectors
using the binarization technique to reduce their memory footprint and speed up the retrieval. VLAD can jointly optimize
dimensionality reduction and indexing algorithms, and thus produce a more compact representation for image retrieval.
Although achieving significant progress in visual retrieval, these works depended heavily on hand-crafted features, which
were not always optimized for particular tasks.

\subsection{Deep representation learning}
Recently, deep representation learning has been successfully applied to various computer vision areas, such as image classification \cite{krizhevsky2012imagenet,ZhuLLL17ivc,Wang2017ICCV}, object detection \cite{girshick2014rich,girshick2015fast,chen2018learning}, pixel-wise image labeling \cite{long2015fully,chen2016disc} and human centric analysis \cite{lin2015deep,chen2016deep}.  We review some recent works that applied deep learning to image retrieval \cite{babenko2014neural,babenko2015aggregating,wang2014learning,paulin2015local,perronnin2015fisher,Jammalamadaka201731}.  Babenko et al. \cite{babenko2014neural} extracted global features for image retrieval from the fully connected layer of CNN pre-trained with an image classification dataset \cite{russakovsky2015imagenet}, and demonstrated fine-tuning the network with annotated target images could boost the retrieval performance.  Their succeeding work \cite{babenko2015aggregating} simply aggregated the average and max pooling of  the last convolutional layer as image representation and achieved better results than using fully convolutional layer. Perronnin and Larlus \cite{perronnin2015fisher} proposed a hybrid image retrieval architecture that combined the strengths of supervised deep representation and unsupervised Fisher Vectors \cite{perronnin2007fisher,perronnin2010improving}. However, these methods extracted deep
features using the CNN trained with a classification objective function, which merely focused on feature learning
and ignored metric learning.

\subsection{Distance learning}

Distance metric learning (DML) plays an important role in many computer vision tasks such as face recognition, person re-identification and image retrieval. Xing et al. \cite{XingNJR02} proposed to learn a distance metric from given examples of similar pairs and demonstrated the learned distance metric can improve clustering performance significantly.
Zheng et al. \cite{ZhengGX11} formulated distance learning as a probabilistic relative distance comparison model to maximize the likelihood in which the examples from a matched pair have smaller distance than those from a mismatched pair. Mignon and Jurie \cite{MignonJ12} proposed a new distance learning method with sparse pairwise constraints.
The success of deep learning also inspired recent works that applied neural network models to address the distance learning problem \cite{LiuMQPZH15,bell2015learning,SongXJS16cvpr}.

The following works are related to our work in spirit of deep metric embedding.
Hu et al. \cite{hu2014discriminative} presented a new discriminative deep metric learning (DDML) method for face verification in the wild. Bell and Bala \cite{bell2015learning} learned deep metric for visual search in interior design using contrastive embedding. \mm{In \cite{RippelPDB15MagnetLoss}, Rippel et al. proposed to maintain an explicit model of the distributions of different classes in representation space and employ it to achieve local discrimination in DML by penalizing class distribution overlap.} Wang et al. \cite{wang2014learning} used a multi-scale CNN trained using triplet embedding for learning fine-grained image similarity directly from the image pixels. Similar ideas were employed in \cite{schroff2015facenet} and \cite{zhang2015bit} for face verification and hash learning, respectively.

Two common embedding methods are closely related to our focus ranking embedding method, i.e., pair \cite{hadsell2006dimensionality,chopra2005learning} and triplet \cite{schroff2015facenet,weinberger2009distance} embedding.
The goal of all the three embedding methods are to push the samples with the same labels close to each other and pull the samples of different labels apart from each other. However, they are quite different in training and optimization.
The pair method builds its energy function on a training set composed of image pairs and optimize to assign small distance to similar pairs and large distance to dissimilar pairs.
The training set of the triplet method is arranged into image triplets (a probe image, a matched one and a reference one). The resulting optimization aims to make the distance of the matched image to the probe image smaller than that of the negative reference one.
However, these two methods do not address the challenging fine-grained fabric image retrieval problem well. The proposed focus ranking method arranges the training set into focus ranking units (see Section \ref{subsec:Optimization}), with much larger number of reference images in contrast to the matched one. Then our optimization penalizes any ranking disorder that negative images in the reference set rank over the matched one. Extensive experimental validation demonstrates that the deep similarity model optimized with our focus ranking embedding method outperforms previous models favorably.

\section{Deep Similarity Model with Focus Ranking}
\label{sec:deeprankingmodel}
%\subsection{Overview}

In this section, we describe the proposed deep focus ranking model in detail. In the training stage, we first arrange
training images into focus ranking units, each of which contains one probe image, one image of the same fabric
(i.e., positive image), and a number of images of different fabrics (i.e., negative images). Then, images in each
focus ranking unit are fed into a CNN to derive features, based on which the ``probe-to-positive'' and
``probe-to-negative'' distances will be computed, respectively. Recall that the probe-to-positive distance is
expected to be constantly smaller than the probe-to-negative distance. Thus, we penalize ranking disorders in
all units. For each pair of input images, the trained CNN will be applied to extract their features separately and
then compute their distance directly in the test stage.

\subsection{Deep metric formulation}

We first present some notations that would be used throughout this article.  Without loss of generalization, we consider the scenario that has $K$ categories of fabrics, in which the $j$-th category has $n_j$ images. Suppose there are $N$ training images $\mathcal{X}=\{x_i|i=1,2,3,...,N\}$, so the training set is $\mathcal{T}=\{(x_i, y_i)|i=1,2,3,...,N\}$, where $x_i$ is the $i$-th training image and $y_i\in\{1,2,3,...,K\}$ denotes the fabric category the image $x_i$ belongs to.
Given an image $x$, there exist some images with the same label, which are the positive ones and denoted by $\mathcal{X^+}$. All the others are considered to be negative ones, denoted by $\mathcal{X^-}$.
A CNN is applied to model the transformation $f(\cdot)$ from a raw image to its feature vector.

Given an image $x$, it is expected that the distances between $f(x)$ and $f(x^+)$ should be smaller than those
between $f(x)$ and $f(x^-)$, for $\forall x^+\in\mathcal{X^+}$ and $\forall x^-\in\mathcal{X^-}$, i.e.,
\begin{equation}
   D(f(x),f(x^+))<D(f(x),f(x^-)),\forall x^+\in\mathcal{X^+},\forall x^-\in\mathcal{X^-},
   \label{eqn:metric_expectation}
\end{equation}
where $D(\cdot,\cdot)$ denotes the distance between two feature vectors. In this paper, we apply $L_2$-norm to measure the distance, i.e., $D(f(x), f(y))=\|f(x)-f(y)\|_2^2$. Intuitively, we penalize the ranking disorder cases that are $D(f(x),f(x^+))>D(f(x),f(x^-))$.  Formally, we apply a 0-1 function to measure the penalty, so the ranking loss of $x$ with respective to $\mathcal{X^+}$ and $\mathcal{X^-}$ can be expressed as
\begin{footnotesize}
\begin{equation}
   loss(x|\mathcal{X^+},\mathcal{X^-})=\sum_{x^+\in\mathcal{X^+}}\sum_{x^-\in\mathcal{X^-}}\textbf{1}(D(f(x),f(x^-))-D(f(x),f(x^+))<0),
    \label{eqn:sample_loss_01}
\end{equation}
\end{footnotesize}
where $\textbf{1}(\cdot)$ is the indictor function whose value is 1 if the expression is true, and 0 otherwise. The loss function can be formulated by summing up the ranking loss over the training set:
$\mathcal{L} = \sum_{x\in\mathcal{X}}loss(x|\mathcal{X^+},\mathcal{X^-})$.

Optimizing the above loss function can minimize the ranking disorder cases and benefit for top ranking the positive samples.
However, the 0-1 loss function is non-differential in optimization, so it is difficult to be jointly optimized with the CNN.
One common solution for this problem is to apply a differential upper-bound function of the 0-1 function.
Inspired by \cite{yun2014ranking}, we replace the $\textbf{1}(x<0)$ by the logistic function \mm{$\sigma(x)=\log_2(1+2^{-x})$}.
The loss function can then be expressed as
\begin{equation}
    \mathcal{L} =\sum_{x\in\mathcal{X}}\sum_{x^+\in\mathcal{X^+}}\sum_{x^-\in\mathcal{X^-}}\sigma(D(f(x),f(x^-))-D(f(x),f(x^+))).
   \label{eqn:total_loss}
\end{equation}

It is worth noting that the proposed model successfully integrates feature extraction and metric learning.
In a traditional method, features are usually hand-crafed, and the selection of a metric is done separately from
feature extraction. In this work, we take advantage of the CNN for feature learning and optimize the two components
jointly.  We will introduce the detailed network architecture and the optimization in the following subsections.

\subsection{Network architecture}
\label{subsec:network_architecture}

\begin{figure*}[!htb]
   \centering
   \includegraphics[width=0.7\linewidth]{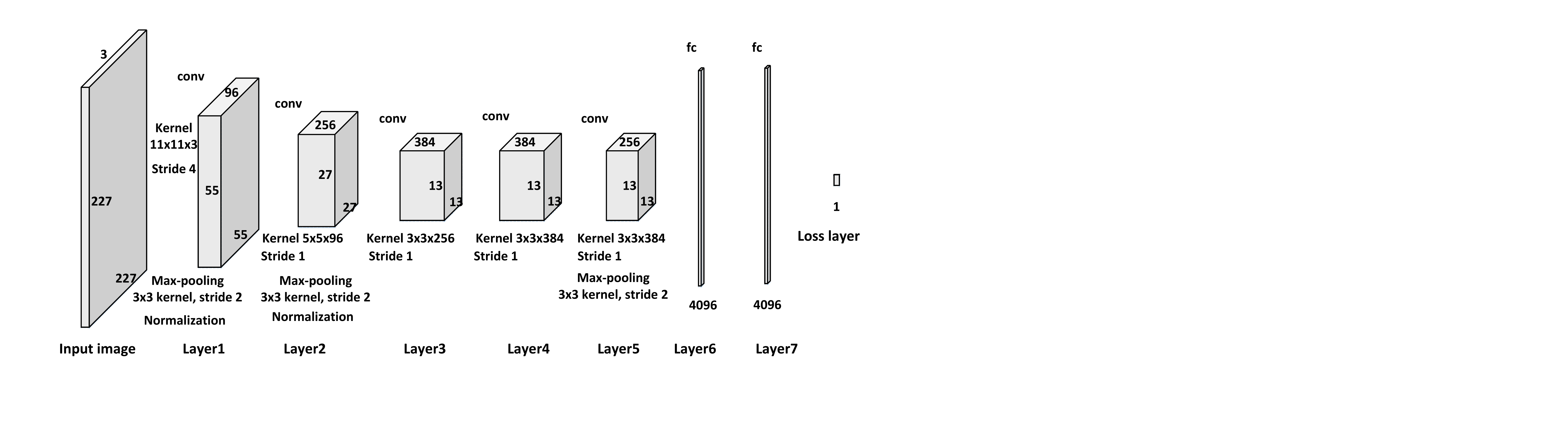} % requires the graphicx package
   \caption{Illustration of our network architecture.}
   \label{fig:network}
\end{figure*}

We build our deep focus ranking model on the popular CNN architecture Alex-Net \cite{krizhevsky2012imagenet}.
The first seven layers of Alex-Net are used as the early layers of our network architecture, and then we add auxiliary structure to fulfill similarity evaluation. Our network architecture is illustrated in Fig. \ref{fig:network}.
Specifically, the first five of the seven layers are convolutional layers and the remaining two are fully connected layers.
The first convolutional layer takes an image of size $227\times 227\times 3$ as input and has 96 kernels of size $11\times 11\times 3$ with a stride of 4 pixels, resulting in 96 feature maps of size $55\times55$. The second convolutional layer has 256 kernels of size $5\times 5\times 96$ and outputs 256 feature maps of size $27\times27$. The output feature maps of the first two convolutional layers are max-pooled ($3\times3$ kernel with stride 2) and normalized before being fed into the next convolutional layer.
The third, fourth, and fifth convolutional layers have 384 kernels of size $3\times 3\times 256$, 384 kernels of size $3\times 3\times 384$, and 256 kernels of size $3\times 3\times 384$, respectively.
The output feature maps of the third and fourth convolutional layers are not intervened by any pooling or normalization, while the output of the fifth convolutional layer is max-pooled ($3\times3$ kernel with stride 2) and results in 256 feature maps of size $6\times6$.
The sixth and seventh fully-connected layers have 4096 neurons each. Moreover, the outputs of all the seven layers are immediately filtered by a rectified linear unit (ReLU) before any pooling or normalization operation.
Finally, we add an auxiliary layer to calculate the loss function (\ref{eqn:total_loss}).

\subsection{Optimization with focus ranking unit generation}\label{subsec:Optimization}
We optimize the proposed model using stochastic gradient descent (SGD) algorithm with momentum \cite{bottou2012stochastic}. It need to feed the whole dataset into the  GPU memory if directly using loss function (\ref{eqn:total_loss}), which is almost intractable for large scale optimization. To tackle this problem, similar to \cite{schroff2015facenet,chen2016deep}, we organize the training data into ranking units to optimize the model. Specifically, for a probe image $x$, we take one positive image $x^+\in\mathcal{X^+}$ and a small subset of $\mathcal{X^-}$, i.e., $U_x\subset{\mathcal{X^-}}$, into consideration. Note that $x^+$, randomly sampled from $\mathcal{X^+}$, is the positive match and $U_x$ randomly sampled from $\mathcal{X^-}$, is the reference set, for probe image $x$. So a focus ranking unit is defined to contain a probe image $x$, a positive match image $x^+$ and the reference set $U_x$. Then the loss can be expressed as
\begin{equation}
    \mathcal{L} =\sum_{x\in\mathcal{X}}\sum_{x^-\in{U_x}}\sigma(D(f(x),f(x^-))-D(f(x),f(x^+))).
   \label{eqn:total_loss_unit}
\end{equation}

\mm{The gradients of the loss with respective to the feature vector can be computed according to the chain rule.
With the aforementioned $\sigma(x)=\log_2(1+2^{-x})$, we first obtain
\begin{scriptsize}
\begin{equation*}
\left\{ \begin{aligned}
  & \frac{\partial \mathcal{L}}{\partial f(x)}=\sum\limits_{x\in \mathcal{X}}{\sum\limits_{{{x}^{-}}\in {{U}_{x}}}{\frac{-\delta (x,{{x}^{+}},{{x}^{-}})}{1+\delta (x,{{x}^{+}},{{x}^{-}})}\left[ \frac{\partial D(f(x),f({{x}^{-}}))}{\partial f(x)}-\frac{\partial D(f(x),f({{x}^{+}}))}{\partial f(x)} \right]}} \\
 & \frac{\partial \mathcal{L}}{\partial f({{x}^{+}})}=\sum\limits_{x\in \mathcal{X}}{\sum\limits_{{{x}^{-}}\in {{U}_{x}}}{\frac{-\delta (x,{{x}^{+}},{{x}^{-}})}{1+\delta (x,{{x}^{+}},{{x}^{-}})}\left[ \frac{\partial D(f(x),f({{x}^{-}}))}{\partial f({{x}^{+}})}-\frac{\partial D(f(x),f({{x}^{+}}))}{\partial f({{x}^{+}})} \right]}} \\
 & \frac{\partial \mathcal{L}}{\partial f({{x}^{-}})}=\sum\limits_{x\in \mathcal{X}}{\sum\limits_{{{x}^{-}}\in {{U}_{x}}}{\frac{-\delta (x,{{x}^{+}},{{x}^{-}})}{1+\delta (x,{{x}^{+}},{{x}^{-}})}\left[ \frac{\partial D(f(x),f({{x}^{-}}))}{\partial f({{x}^{-}})}-\frac{\partial D(f(x),f({{x}^{+}}))}{\partial f({{x}^{-}})} \right]}}
\end{aligned} \right.
\end{equation*}
\end{scriptsize}
where
\begin{equation} % requires amsmath; align* for no eq. number
   \delta(x, x^+, x^-)=2^{-\left[D(f(x),f(x^-))-D(f(x),f(x^+))\right]}.
\end{equation}
}

\mm{Remembering that $D(\cdot,\cdot)$ is the $L_2$ norm, then we can obtain the derivatives as follows:
\begin{equation}
\left\{ \begin{aligned}
  & \frac{\partial \mathcal{L}}{\partial f(x)}=2\sum\limits_{x\in \mathcal{X}}{\sum\limits_{{{x}^{-}}\in {{U}_{x}}}{\frac{-\delta (x,{{x}^{+}},{{x}^{-}})}{1+\delta (x,{{x}^{+}},{{x}^{-}})}\left[f({{x}^{+}})-f({{x}^{-}})\right]}} \\
 & \frac{\partial \mathcal{L}}{\partial f({{x}^{+}})}=2\sum\limits_{x\in \mathcal{X}}{\sum\limits_{{{x}^{-}}\in {{U}_{x}}}{\frac{-\delta (x,{{x}^{+}},{{x}^{-}})}{1+\delta (x,{{x}^{+}},{{x}^{-}})}\left[f(x)-f({{x}^{+}})\right]}} \\
 & \frac{\partial \mathcal{L}}{\partial f({{x}^{-}})}=2\sum\limits_{x\in \mathcal{X}}{\sum\limits_{{{x}^{-}}\in {{U}_{x}}}{\frac{-\delta (x,{{x}^{+}},{{x}^{-}})}{1+\delta (x,{{x}^{+}},{{x}^{-}})}\left[f({{x}^{-}})-f(x)\right]}}
\end{aligned} \right.
\end{equation}
}

We employ the unit sampling generation to optimize the model for the following serval reasons. First, taking a random subset in each iteration and applying sufficient iterations is approximately equivalent to using the whole reference set. Second, it is more efficient and practical to take a subset for training, especially for large-scale learning. Third, training the model with a random subset introduces a high degree of diversity, which is of great importance to learning.

\section{Fabric Image Retrieval Dataset (FIRD)}
\label{sec:dataset}

\begin{figure*}[!htb]
   \centering
   \includegraphics[width=0.6\linewidth]{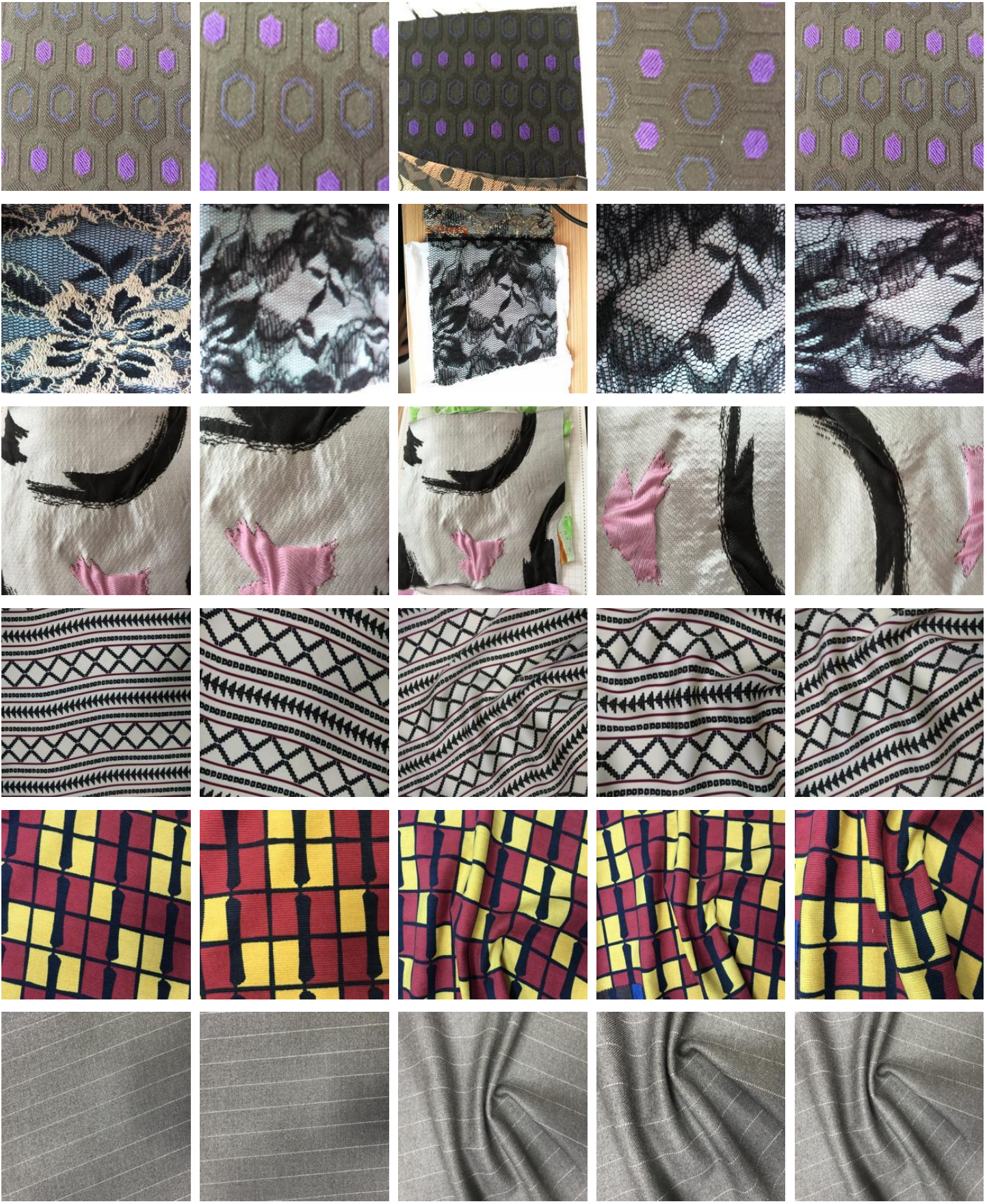} % requires the graphicx package
   \caption{Samples from the FIRD dataset under different photographing conditions.}
   \label{fig:FIRD}
\end{figure*}

To the best of our knowledge, there are not public datasets for the fabric image retrieval problem.
Since fabric image retrieval benefits a wide range of application, e.g., large-scale clothing searching and
online shopping, we construct a large-scale fabric image retrieval dataset (FIRD).
Specifically, we collect \mm{4,300 fabrics}, and take different photos for each fabric. In order to imitate real application
scenarios, we vary the collecting conditions from the following perspectives.

\noindent\textbf{Photographing device:} Without loss of generalization, we use cell phones for collecting images as most of clothing searching and online shopping  applications are used in cell phones. We use varied cell phones, from Android to iOS, from low- to high-end, to increase the diversity of collected images.

\noindent\textbf{Photographing circumstance:} We take photos of a fabric in varied circumstances, including different scenes (indoor and outdoor), time (day and night), and illumination conditions.

\noindent\textbf{Photographing view:} We change viewing conditions as follows: (1) we rotate the fabric to collect images with different angles; (2) we take photos in near and far views to collect images with different scales; \mm{(3) we wrinkle the fabric to collect fabric images with various degrees of wrinkle.}

\mm{Figure \ref{fig:FIRD} illustrates some samples from the FIRD dataset. It is a challenging dataset due to the complex variations in illumination, rotation, scale and wrinkle. All the 4,300 fabrics are incorporated in the dataset and, for each fabric, 5--10 images are collected under different above-mentioned conditions. The images of the same fabric are annotated by the same label (the $i$th fabric labelled $i$). We randomly select half of the 4,300 fabrics as training set and the rest as test set. So both training and test sets contain 2,150 categories of fabric. During testing, we randomly select 2/5 images of each category in the test set as query images, and arrange the rest images as a retrieval set. Given a query image, we aim to find the ones with the same label.}

\section{Experiments}
\label{sec:experiment}
In this section, we will present experimental comparions between the proposal method and existing metric learning methods.
We will also present intra-model comparisons to analyze the contribution of focus ranking unit generation.

\subsection{Evaluation protocols}
We utilize the following measures to give an extensive evaluation and analysis of the tested methods' performance.

\mm{\textbf{Recall@K.}\; One of the critical measures for evaluating the retrieval performance is the recall at particular ranks, i.e., Recall@K \cite{SongXJS16cvpr,jegou2011product}. In this measure, given a query image, the images with top $K$ smallest distances to it are retrieved, and Recall@K is the ratio of retrieved relevant images over the total relevant images. The Recall@K on the dataset is then computed by averaging the scores of all query images.}

\mm{\textbf{mAP.}\; Another commonly used measure is the mean average precision (mAP) \cite{PhilbinCISZ07cvpr}. For each query image, a precision/recall curve can be obtained, and its average precision (the area under the curve) is computed. The mAP is obtained as the mean over all the queries.}

\mm{\textbf{F1 and NMI scores.}\; To further demonstrate the superiority of the proposed model, we also apply F1 and NMI scores \cite{schutze2008introduction} to evaluate the performance. The Normalized Mutual Information (NMI) score is calculated by conducting clustering experiments on the test set with the learned features.}

\subsection{Compared methods and implementation details}

\mm{We compare our model with five methods, i.e., Alex baseline, Softmax, pair, triplet and Magnet. Among them, the Alex baseline method applies the Alex-net model \cite{krizhevsky2012imagenet}, which is trained on the ImageNet classification dataset \cite{russakovsky2015imagenet} using the Caffe framework \cite{jia2014caffe}, and extracts the features of the first fully connected layer (i.e., fc6), so that the best performance can be ensured, as suggested in \cite{babenko2014neural}. The Softmax method uses the Alex-net model pre-trained on ImageNet and then fine-tunes it on our FIRD training set. Afterwards, the feature extracted on the fc6 layer is used for evaluation. We also note that, for fair comparison, all the other tested methods are based on the Alex-net model and use the feature from the fc6 layer.}

\mm{The Magnet model applies the Magnet loss \cite{RippelPDB15MagnetLoss}. It is a recent deep metric embedding method and sculpts the feature representation by identifying and handling intra-class variation and inter-class similarity. We use its publicly available TensorFlow code\footnote{https://github.com/pumpikano/tf-magnet-loss} and the Alex-net tensorflow model\footnote{http://www.cs.toronto.edu/\~{}guerzhoy/tf\_alexnet/} pre-trained on ImageNet to train on our FIRD dataset. We initialize the learning rate as 0.01, with momentum of 0.9, weight decay of 0.0005, and batch size of 80 (each batch randomly selects 16 classes and 5 images for each class).}

\mm{The two common deep metric embedding methods (i.e., pair and triplet) and the proposed method are implemented in the Caffe framework.
These models are initialized with the parameters of the Alex-net model pre-trained on ImageNet and fine-tuned on our training set using the SGD method with back-propagation. For pair and triplet models, we initialize the learning rate as 0.001, with momentum of 0.9, weight decay of 0.0005, and batch sizes of 80 and 120 respectively. For our focus ranking model, we fix the positive and negative ratio as 1:32 in the focus ranking unit, which can well approximate the proposed model.}

\mm{During training, the images are resized to $256\times256$. In addition, the images are randomly cropped and randomly mirrored horizontally for data augmentation. During inference, all images are resized to $256\times256$ and cropped $227\times227$ at the image center for feature extraction. All experiments were tested on a single NVIDIA Tesla K40 GPU.}

\subsection{Comparison results}
Note that the original feature dimensions of all models are 4,096, and they are too large for an efficient retrieval system. Therefore, we follow Babenko et al. \cite{babenko2014neural} to apply the Principal Component Analysis (PCA) \cite{jolliffe2002principal} to reduce the dimension to smaller ones (i.e., 512, 64, 16, 8), and perform comparisons on various feature dimensions.

\begin{figure*}[!htb]
\centering
\subfigure[]{
%\label{fig:subfig1_1} %% label for first subfigure
\includegraphics[width=0.33\linewidth]{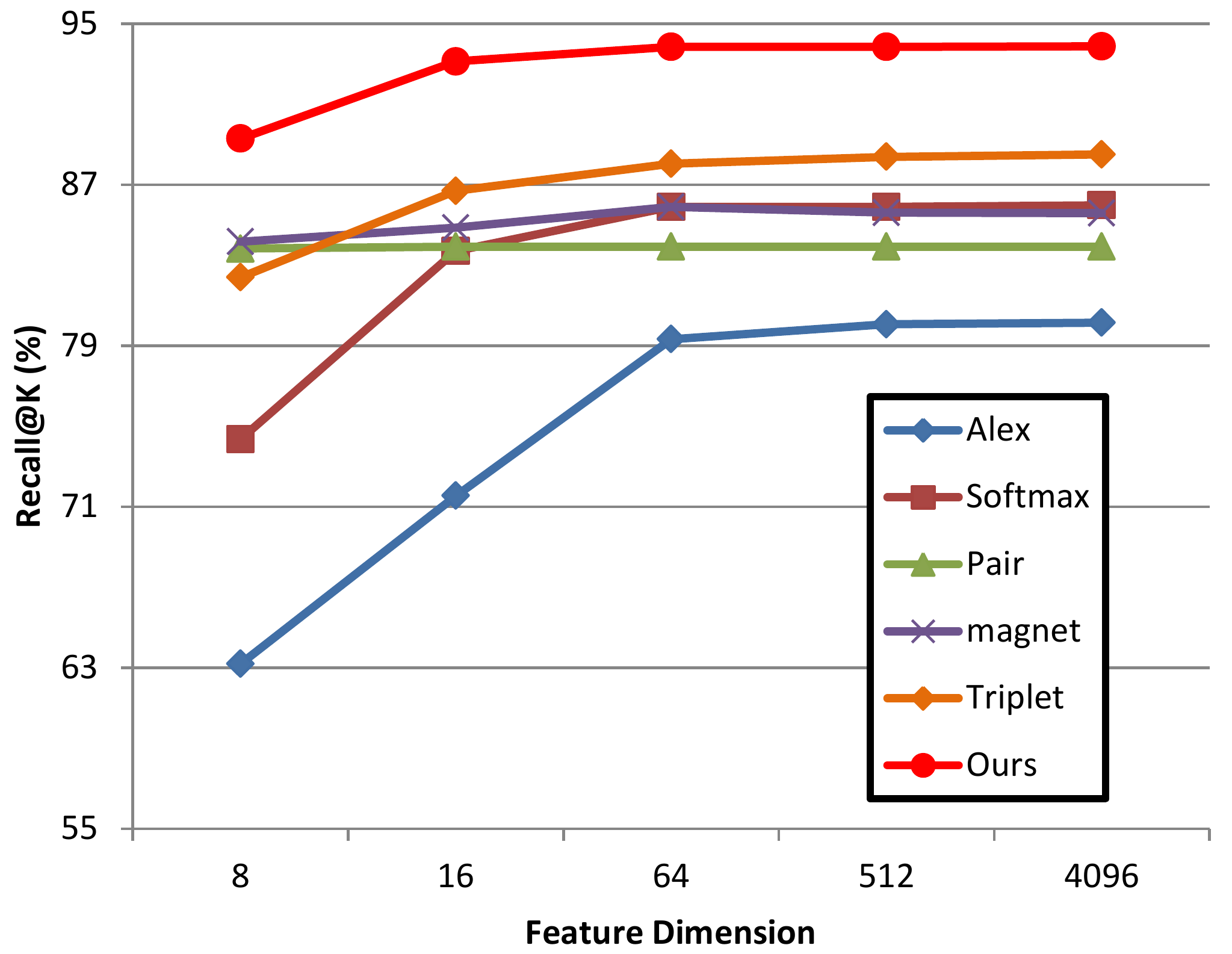}}
\subfigure[]{
%\label{fig:subfig1_2} %% label for second subfigure
\includegraphics[width=0.33\linewidth]{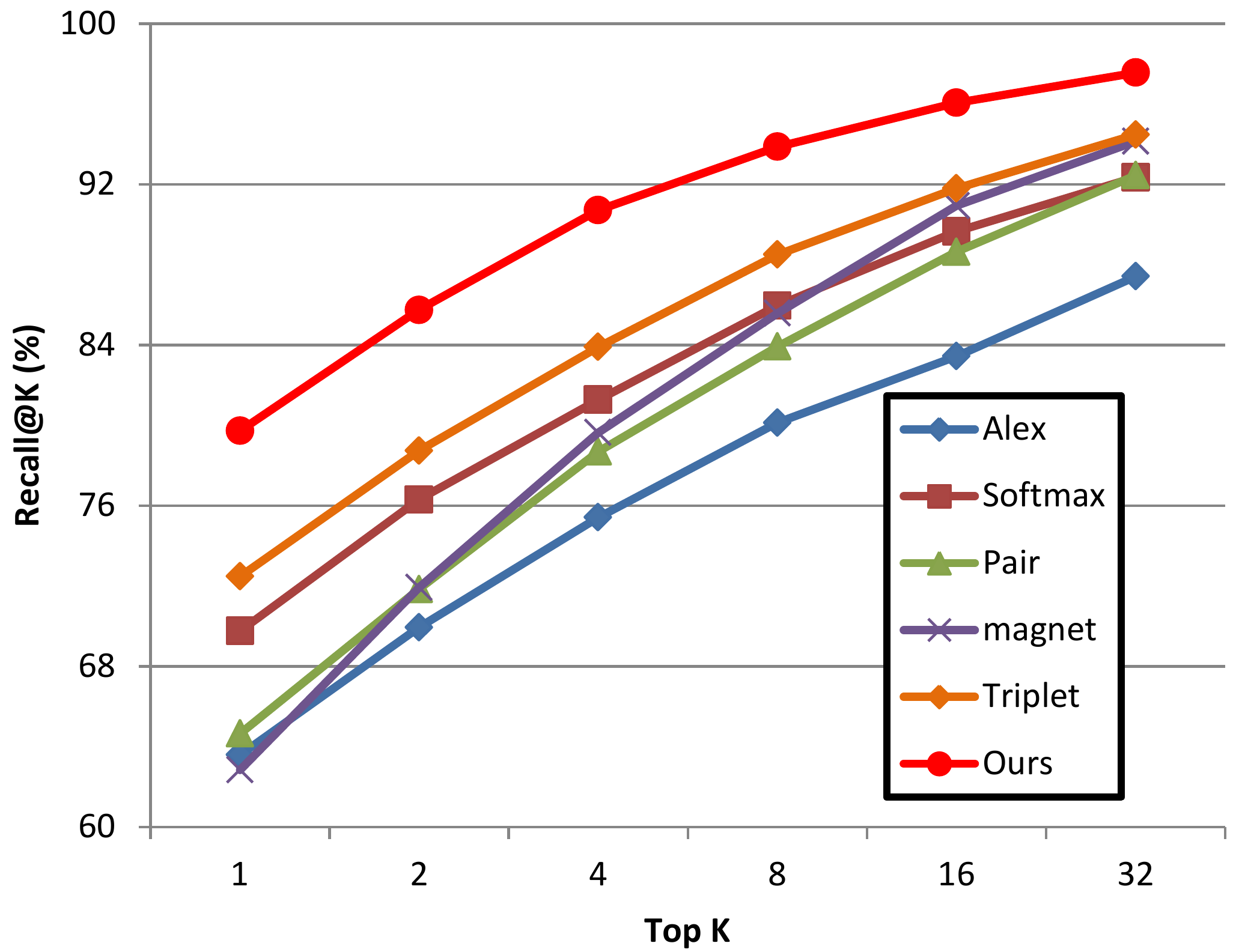}}
\caption{Comparison on the FIRD dataset using the Recall@K measure. (a) Recall@K vs. feature dimension with K=16. (b) Recall@K vs. K with the feature dimension of 256. Best viewed in color.}
\label{fig:comparison-Recall@K}
\end{figure*}

\begin{figure*}[!htb]
\centering
\includegraphics[width=0.33\linewidth]{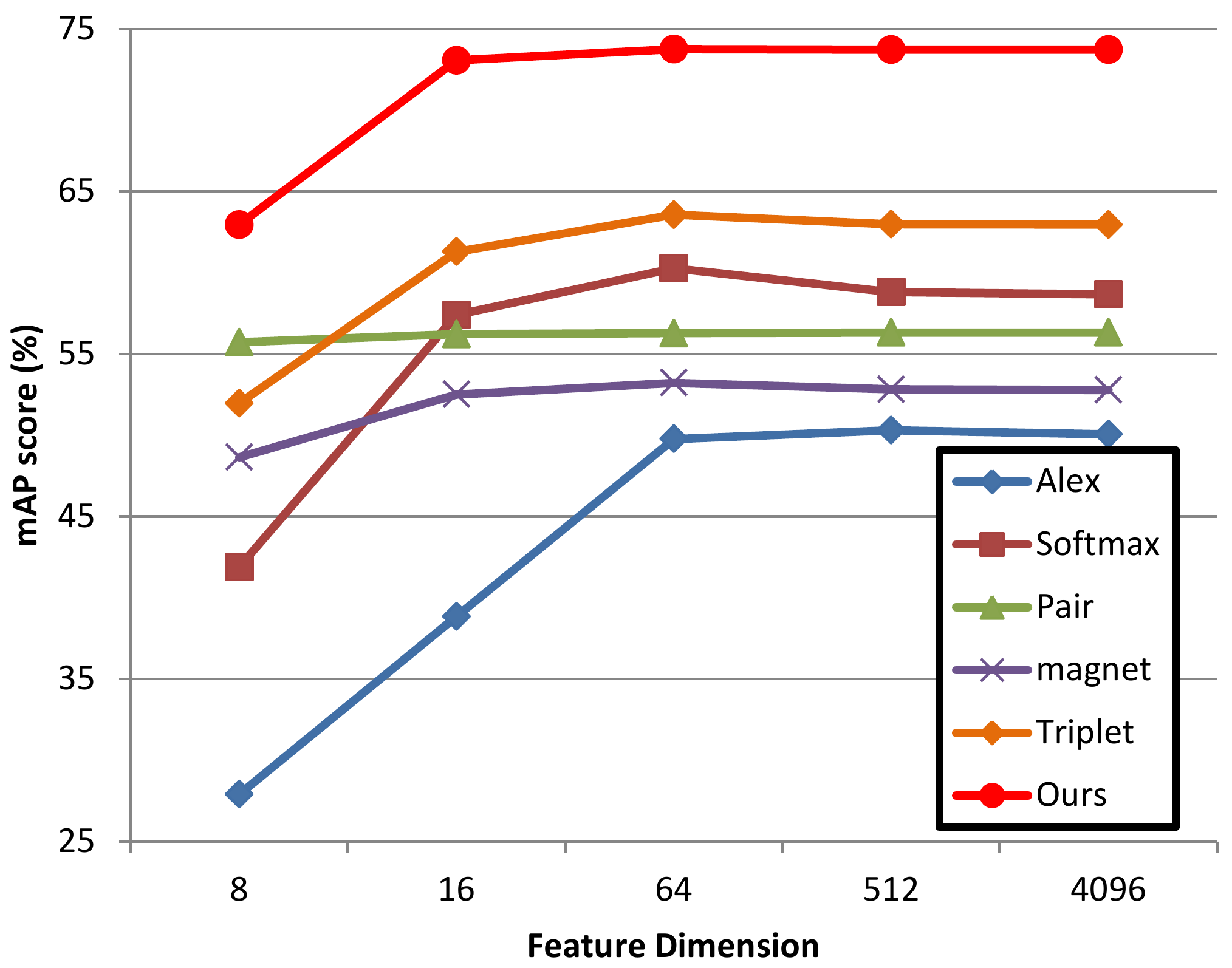}
\caption{Comparison on the FIRD dataset using the mAP measure. Best viewed in color.}
\label{fig:comparison-mAP}
\end{figure*}

\begin{figure*}[!htb]
\centering
\subfigure[]{
%\label{fig:subfig1_1} %% label for first subfigure
\includegraphics[width=0.33\linewidth]{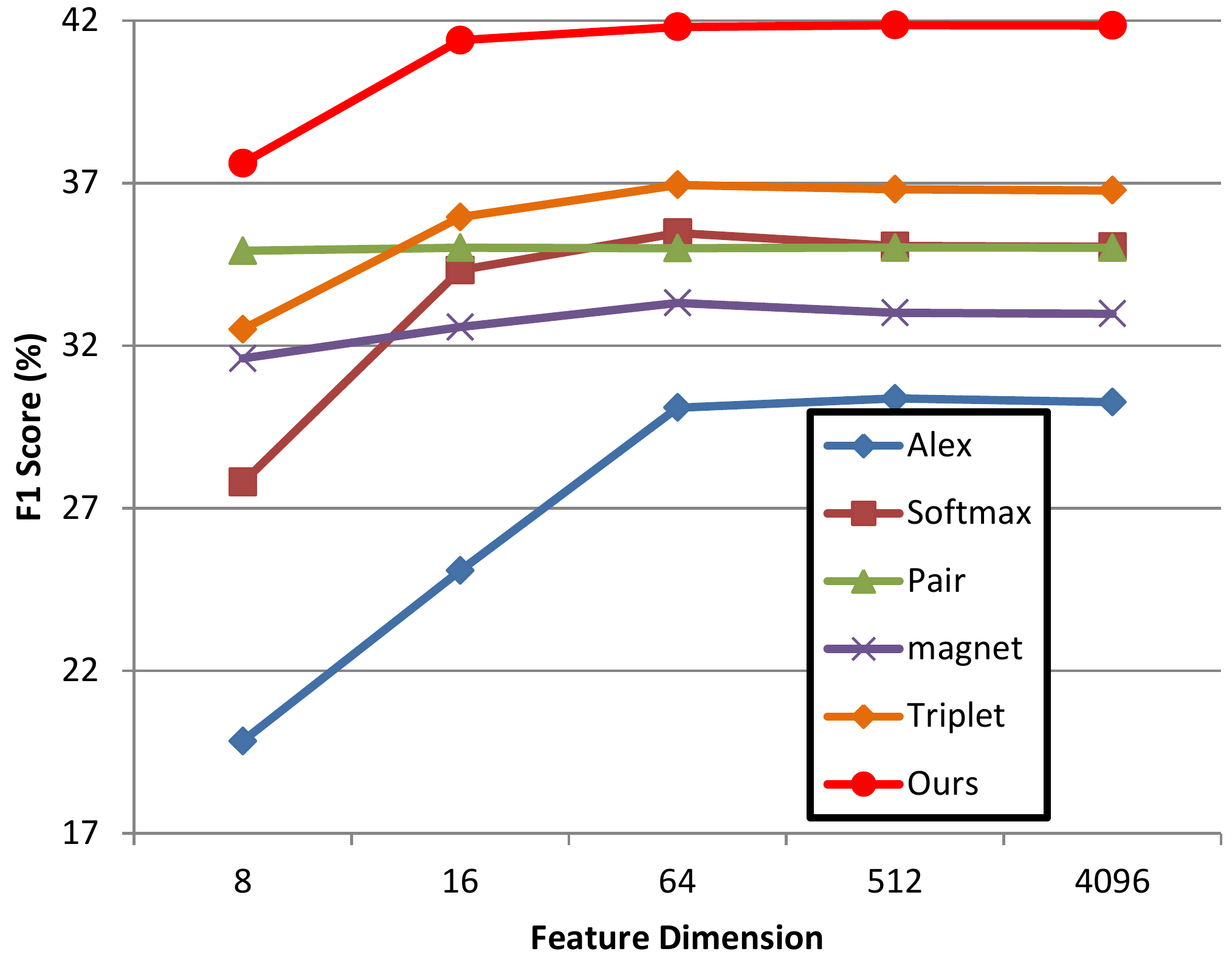}}
\subfigure[]{
%\label{fig:subfig1_2} %% label for second subfigure
\includegraphics[width=0.33\linewidth]{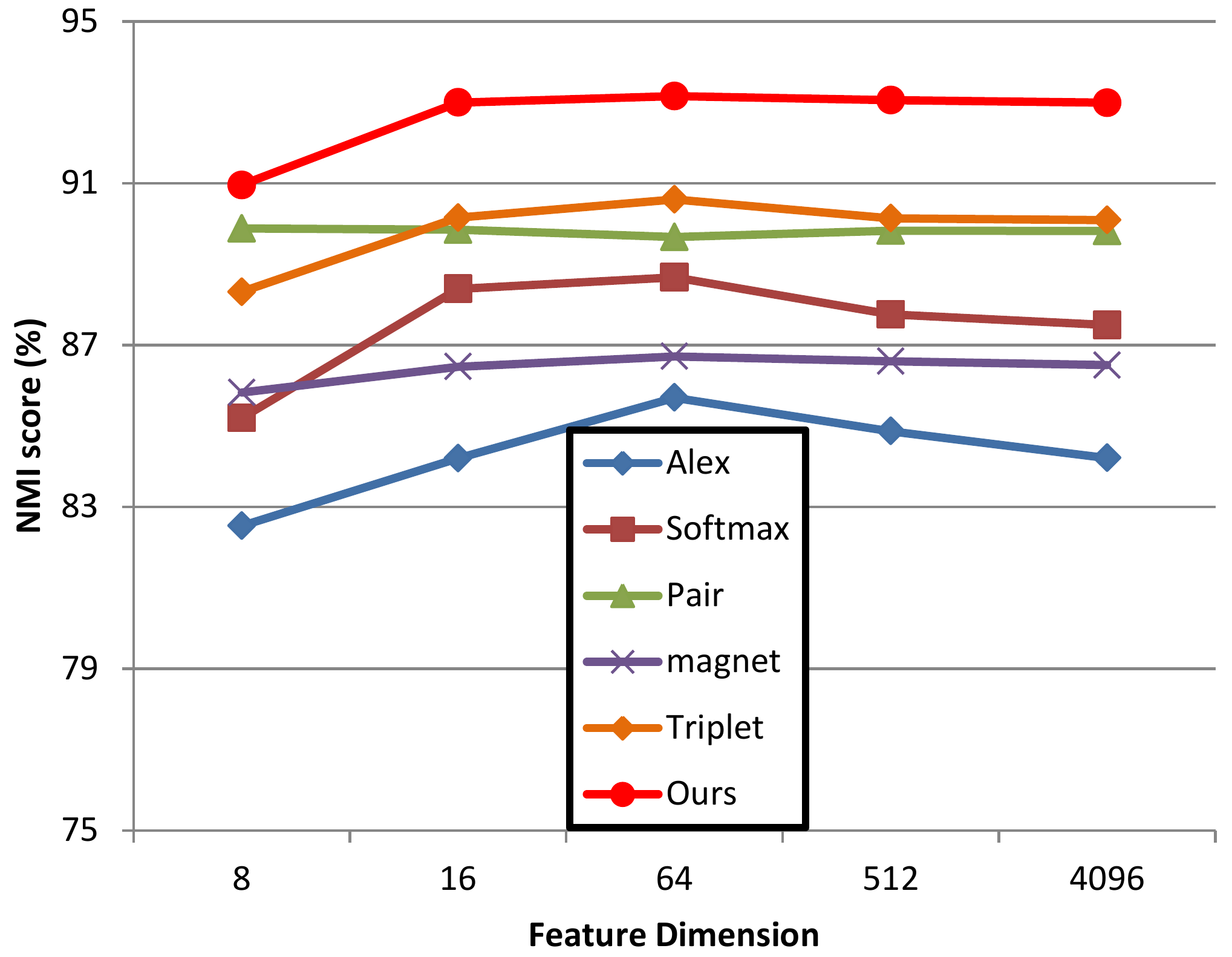}}
\caption{Comparison on the FIRD dataset with F1 and NMI scores. (a) F1 score vs. feature dimension. (b) NMI score vs. feature dimension. Best viewed in color.}
\label{fig:comparison-F1-NMI}
\end{figure*}

\begin{figure*}[!htb]
   \centering
   \includegraphics[width=0.7\linewidth]{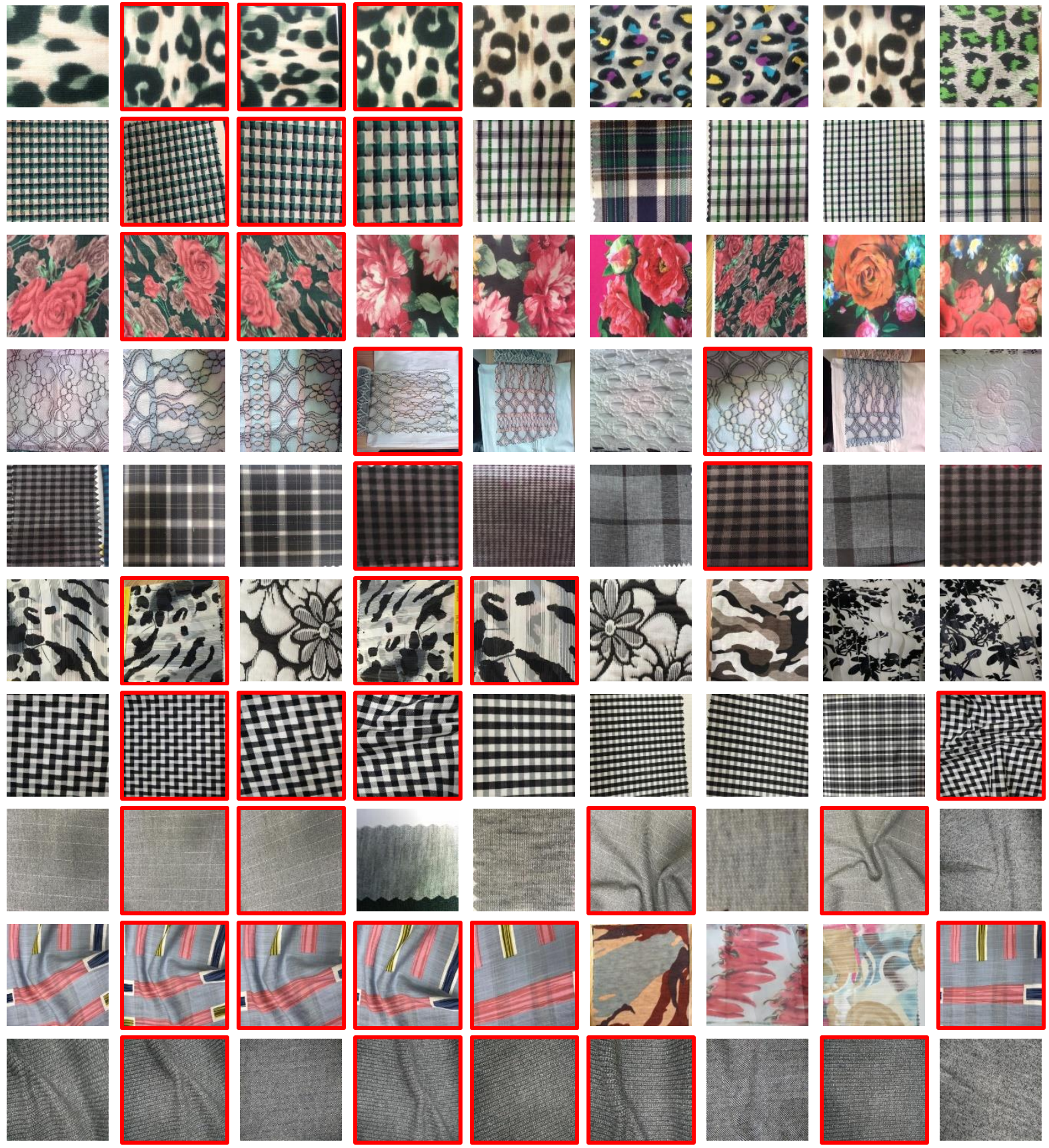} % requires the graphicx package
   \caption{Retrieval results of query images randomly selected from the FIRD dataset. The first column are the query images, and the rest are the retrieved ones, arranged according to the distance to the corresponding query images. The retrieved images of the same class to the query image are marked in red boxes.}
   \label{fig:visualization_random}
\end{figure*}

\begin{figure*}[!htb]
   \centering
   \includegraphics[width=0.7\linewidth]{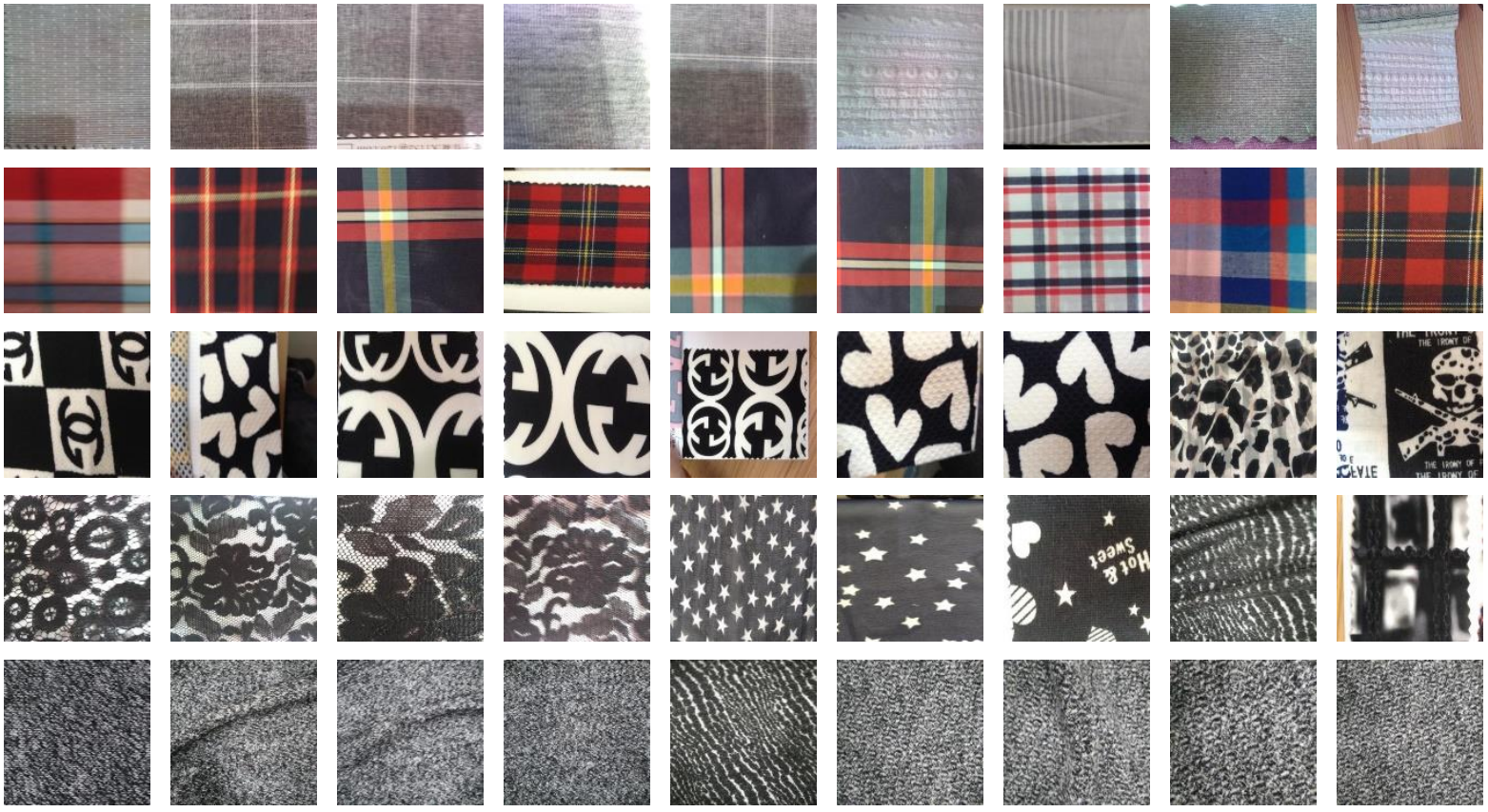} % requires the graphicx package
   \caption{Retrieval results of failure query images selected from the FIRD dataset. The first column are the query images, and the rest are the retrieved ones, arranged according to the distance to the corresponding query images.}
   \label{fig:visualization_failure}
\end{figure*}

\begin{figure*}[!t]
\centering
\subfigure[]{
%\label{fig:subfig1_1} %% label for first subfigure
\includegraphics[width=0.33\linewidth]{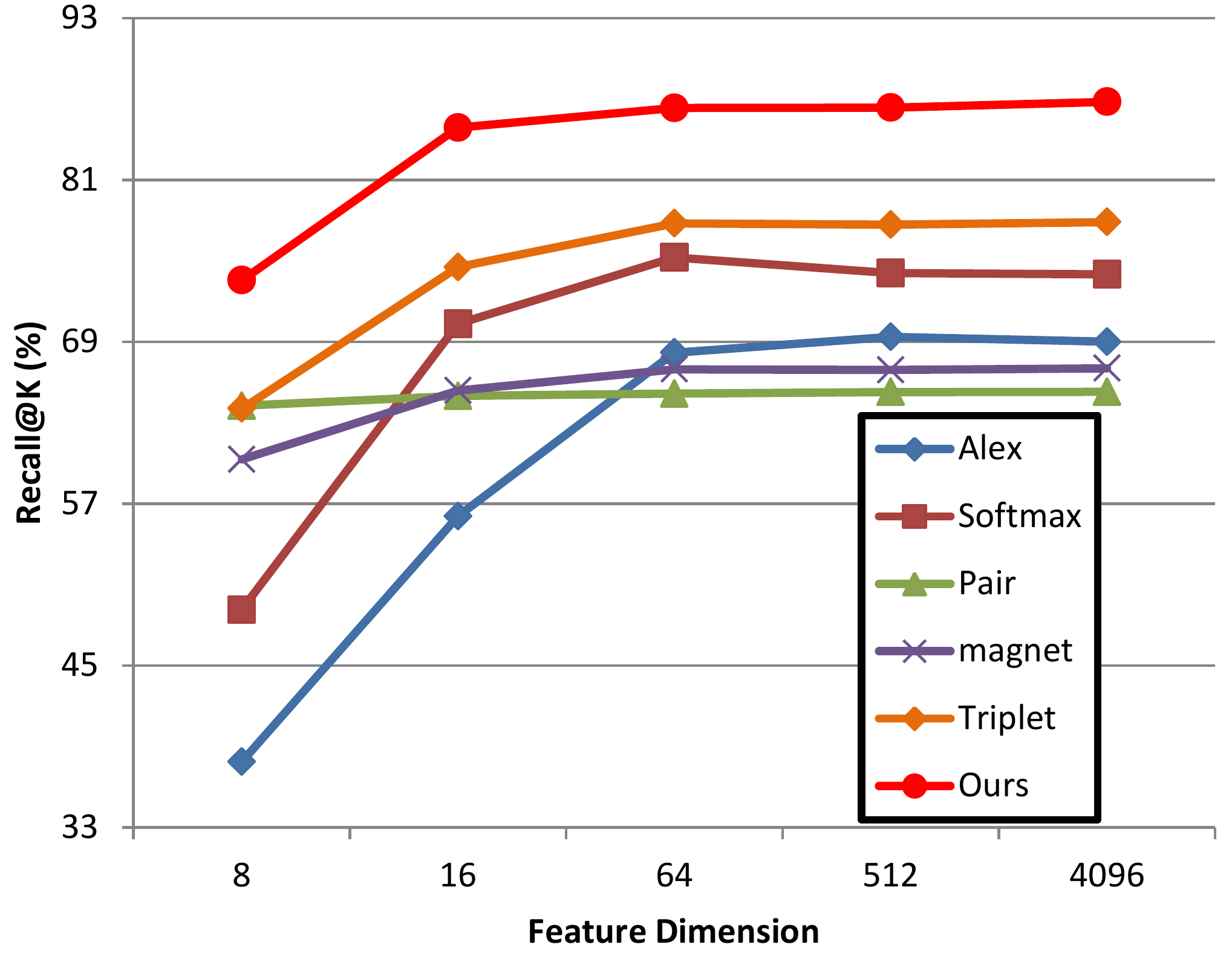}}
\subfigure[]{
%\label{fig:subfig1_2} %% label for second subfigure
\includegraphics[width=0.33\linewidth]{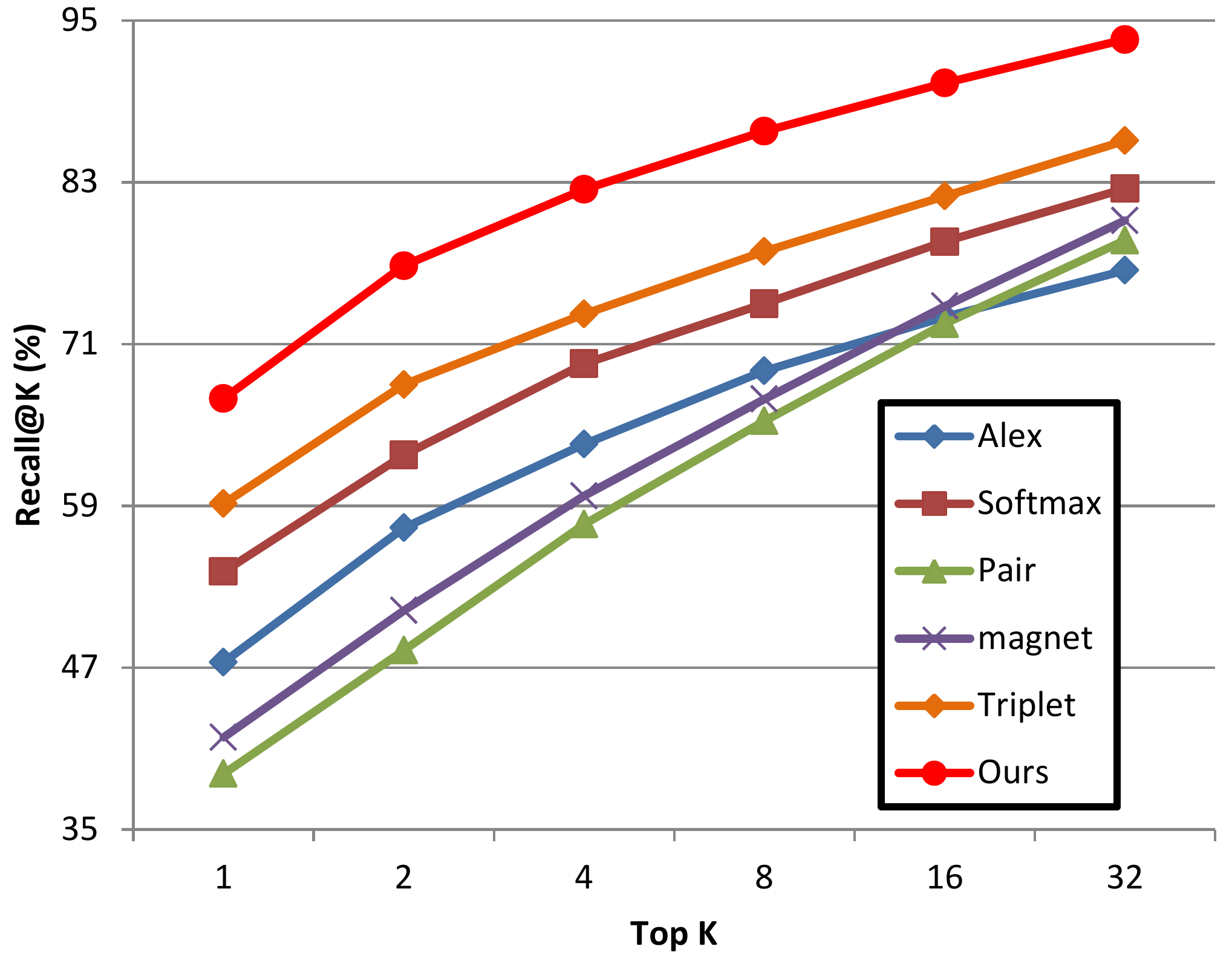}}
\caption{Comparison on the FIRD dataset with additional 100,000 fabric images. (a) Recall@K vs. feature dimension with K=16. (b) Recall@K vs. K with the feature dimension of 256.  Best viewed in color.}
\label{fig:comparison_result_w_add_data}
\end{figure*}

\begin{figure*}[!t]
\centering
\subfigure[]{
%\label{fig:subfig1_2} %% label for second subfigure
\includegraphics[width=0.33\linewidth]{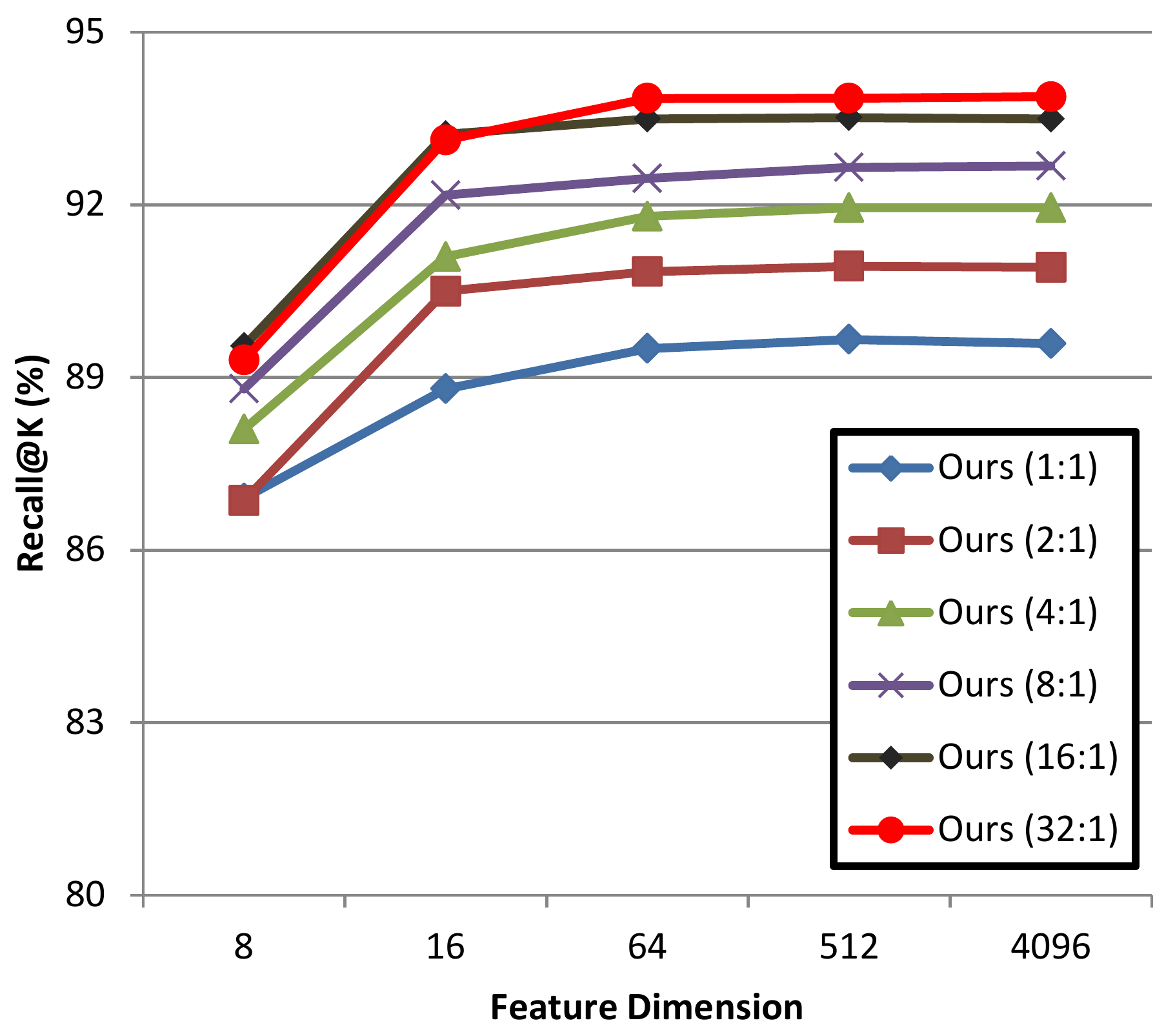}}
\subfigure[]{
%\label{fig:subfig1_1} %% label for first subfigure
\includegraphics[width=0.33\linewidth]{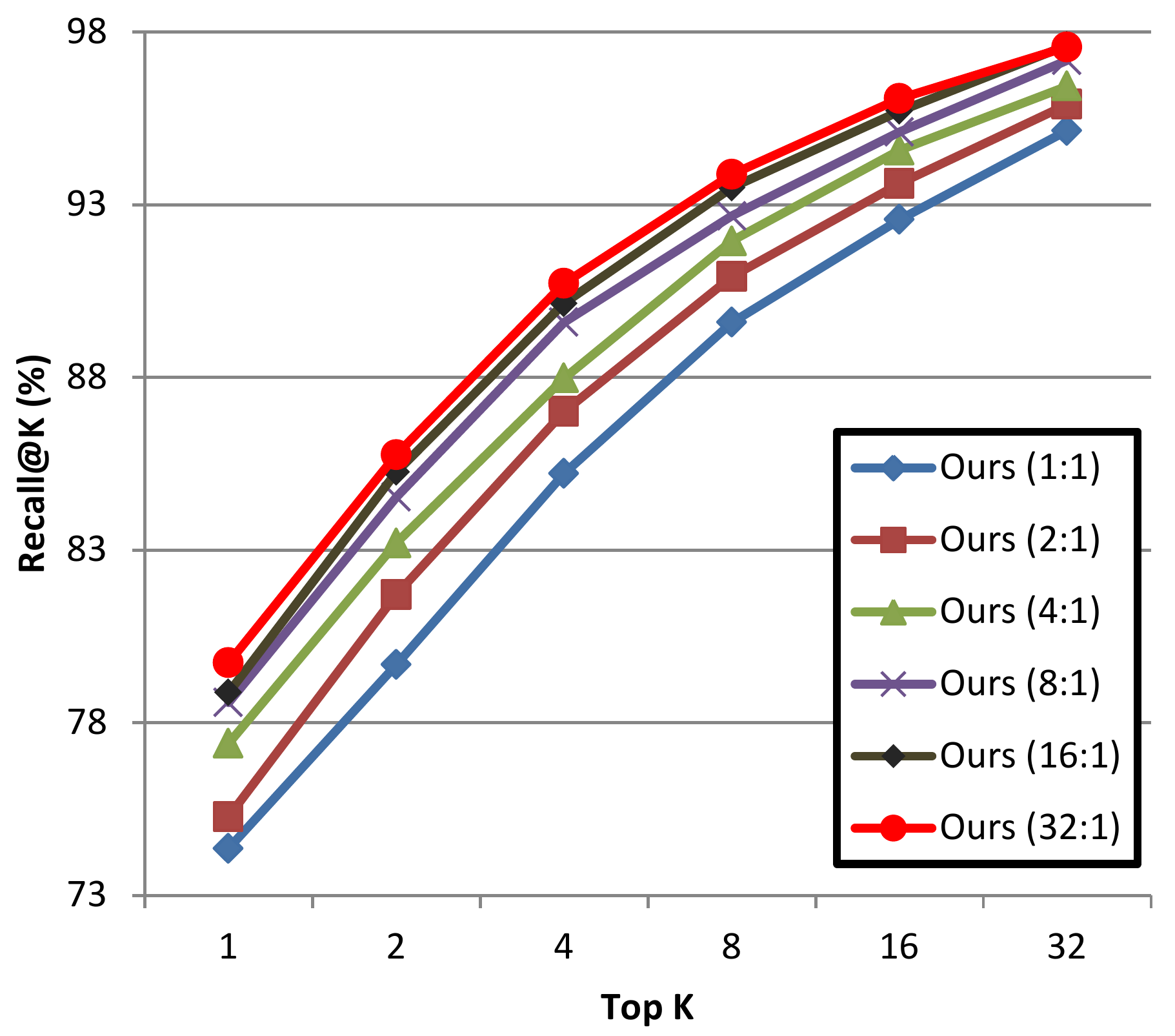}}
\caption{Analysis of focus ranking unit generation. (a) Recall@K vs. feature dimension with K=16. (b) Recall@K vs. K with the feature dimension of 256. Best viewed in color.}
\label{fig:rank_unit}
\end{figure*}

\mm{We first perform comparisons on the FIRD test set. For each fabric, we randomly select 2/5 images of each category as queries and use the rest as retrieval ones, making up a retrieval set of about 7,500 images. The comparison results of Recall@K are presented in Figure \ref{fig:comparison-Recall@K}. It can be observed that all the other tested models perform well beyond the Alex baseline one.
In addition, Figure \ref{fig:comparison-Recall@K}(a) reveals that, as the feature dimension is reduced from 4,096 to 16, the recall rates of the four metric embedding models are nearly fixed but the Alex baseline and Softmax models suffer from a severe drop, suggesting that models with metric embedding can generate more compact and discriminative features.
Among the four metric embedding models, our model can still outperform the other three by a large margin.
For example, with the feature dimension of 4096, although the pair, Magnet and triplet models can achieve the high recall rates of 83.9\%, 85.6\% and 88.5\% respectively, our model further improves them to 93.9\%.
Figure \ref{fig:comparison-Recall@K}(b) reports the Recall@K scores when changing the number of retrieved images K from 1 to 32 and fixing the the feature dimension as 256.  Since more relevant images can be retrieved when K increases, so the scores of all the tested methods will also monotonously increase. The figure shows that our method has a superior performance over the other methods at each value of K.}

\mm{Figure \ref{fig:comparison-mAP} shows the comparison results measured in mAP with respect to the feature dimension. It reveals again the other tested models all perform better than the Alex baseline one. It should be noted that the Softmax model with fine-tuning on the FIRD training set has significant performance gain from the Alex one. The evaluation results in mAP further demonstrate the superiority of our model.
It suggests that our model can perform much better than the existing ones in fabric image retrieval systems.}

\mm{We also present the comparison results of F1 and NMI scores in Figure \ref{fig:comparison-F1-NMI}. The F1 score takes both accuracy and recall into consideration, and the results depicted in Figure \ref{fig:comparison-F1-NMI}(a) give further approval in the effectiveness of our model. We also conduct clustering experiments on the FIRD test set using the features extracted by all the tested models, and Figure \ref{fig:comparison-F1-NMI}(b) reports the comparison results measured in NMI. Our model performs much better than all the others, suggesting that it can also achieve much better clustering quality.}

\mm{In summary, observed from the quantitative performance of all the tested models in different measures, the effectiveness of the proposed method is sufficiently verified. We also observe that the pair model's performance is the least sensitive to feature dimension changes, while the triplet model's performance ranks second in most measures. The Softmax model exhibits impressive performance improvement from the Alex baseline model and can outperform the pair model in some measures, which shows that fine-tuning on the target dataset can boost the baseline model's performance. For the Magnet model, it does not work quite well in our experiments. We should note that it is implemented in TensorFlow, different from the others, and better explicit models of distributions of classes and more careful fine-tuning may boost its performance further.}

To give a qualitative analysis of our method, Figure \ref{fig:visualization_random} presents the top 8 retrieved images of some random query examples and Figure \ref{fig:visualization_failure} presents those of some failure cases. We find that, in most cases, our model can successfully retrieve the images belonging to the same fabric, despite of \mm{the complex variations in illumination, rotation, scale, and wrinkle}. Moreover, most images of the same fabric rank top, and other retrieved ones also have very similar textural pattern with the query ones, which is also very essential for real-world systems. On the other hand, for the failure examples, most retrieved images come from the fabrics with subtle differences to the query.

In order to better mimic the real-world scenarios, we further do experiments with a much larger dataset. We collect 100,000 fabrics and take one image for each fabric under the varied photographing conditions, making up a dataset of 100,000 fabric images. \mm{The original retrieval set of about 7,500 images and these 100,000 images are mixed to construct a new retrieval set.} The retrieval results using the new retrieval set are presented in Figure \ref{fig:comparison_result_w_add_data} with the Recall@K measure. It shows that our model again suppresses all other methods by an even larger margin, suggesting our model is more suitable for real-world large scale fabric retrieval systems.

\subsection{Analysis of focus ranking unit generation}

In this work, we generate focus ranking units by sampling to approximate our ranking model in the process of model optimization, and the negative-positive ratio is a very critical parameter. In this section, we analyze the ranking unit generation algorithm by evaluating the performance when training the model with the negative-positive ratio of 1:1, 2:1, 4:1, 8:1, 16:1 and 32:1 on the FIRD dataset. Figure \ref{fig:rank_unit} presents the quantitative evaluation results measured in Recall@K. It shows that there are obvious improvement consistently when increasing the ratio from 1:1 to 16:1. However, the improvement of recall rate when increasing the ratio from 16:1 to 32:1 is negligible, and increasing the ratio further can hardly obtain more improvement on the recall rate. These experimental results show that increasing the ratio may boost the performance by quite a large margin, but it becomes near saturation when the ratio reaches 32:1. It also suggests that a ratio of 32:1 can well approximate our ranking model.

\section{Conclusion}
\label{sec:conclusion}
In this paper, we propose a novel focus ranking embedding method, which aims to rank similar examples top over all dissimilar ones. The focus ranking method can easily be unified into a deep CNN to jointly learn an image representation and a metric. At the training stage, we apply focus ranking unit generation to approximate the proposed model for efficient training, and at the inference stage, a test image can be directly fed into the learnt CNN to extract the features in one shot. Extensive experimental results suggest that our deep focus ranking model shows superior performance over existing metric embedding methods.

\section*{Acknowledgment}
This work was supported in part by the National Natural Science Foundation of China (61379112, 61402120, 61672547, 61472455, 61370186). The authors would like to thank Shenzhen Micro Vision Technology Co., Ltd. for providing the large-scale dataset of 100,000 fabric images for evaluation.

%\section*{References}

\bibliography{mybibfile}
\end{document}